%% file: acl_latex.tex
\pdfoutput=1

\documentclass[11pt]{article}

\usepackage[]{acl}

\usepackage{times}
\usepackage{latexsym}
\usepackage{tikz}
\usepackage{amsmath}
\usetikzlibrary{arrows, arrows.meta, positioning, calc}

\usepackage[T1]{fontenc}

\usepackage[utf8]{inputenc}

\usepackage{microtype}

\DeclareMathOperator*{\argmax}{arg\,max} 

\makeatletter
\newcommand{\specificthanks}[1]{\@fnsymbol{#1}}
\makeatother

\title{Persona-Knowledge Dialogue Multi-Context Retrieval and Enhanced Decoding Methods}

\author{Min Sik Oh\thanks{\hspace{2mm}co-first authors.} \thanks{\hspace{2mm}work done outside Amazon.} \\
  Alexa AI \\
  \texttt{ohtrent@amazon.com} \\\And
  Min Sang Kim\textsuperscript{\specificthanks{1}} \\
  Kakao Enterprise \\
  \texttt{lucas.ai@kakaoenterprise.com} \\}

\begin{document}
\maketitle
\begin{abstract}
Persona and Knowledge dual context open-domain chat is a novel dialogue generation task introduced recently~\cite{pkchat2022}. While Persona and Knowledge is each interesting context of open-domain dialogue, the combination of both has not been well studied. We tackle Persona-Knowledge identification and response generation tasks in this paper. We design an informed data augmentation strategy that is compatible with neural Q\&A retrieval models. With the augmented data, we perform permutative Persona-Knowledge evaluation and successive Persona search fine-tuning. Furthermore, we perform dialogue generation with various decoding techniques and illustrate crucial elements. We achieve SOTA across official metrics with $93.99\%$ Grounding accuracy average and $23.62$ SacreBLEU score.
\end{abstract}

\input{introduction}

\input{related}

\input{methodology}

\input{experiment_setup}

\input{results}

\input{conclusion}

\bibliography{anthology,custom}
\bibliographystyle{acl_natbib}

\appendix

\section{Appendix}
\label{sec:appendix}
While we achieve strong performance increase without any training of generative model, we find that our experimental results do not fully agree with existing methods introduced in~\cite{roller2020recipes} and~\cite{wu2016google}. Robust decoding method applicable to multiple open-domain dialogue domains could be found. We leave this question to future studies.

\subsection{The effect of the minimum length}

Performance with different decoding minimum lengths. Other parameters are same as ours (5 epoch).
\begin{table}[h]
    \centering
    \begin{tabular}{|c|cccc|}
    \hline
     min. length & 5 & 10 & 20 & 40\\
     \hline
     BLEU & \textbf{16.18} & 16.17 & 15.82 & 13.86 \\
     \hline
    \end{tabular}  
    \caption{Different minimum lengths result.}
\end{table}

\subsection{The effect of the length normalization}
Performance with different alpha coefficients of the length normalization. Other parameters are the same as ours (5 epoch).
\begin{table}[h]
    \begin{tabular}[width=0.5]{|c|ccccc|}
    \hline
    alpha & 0.2 & 0.4 & 0.6 & 0.8 & 1.0 \\
    \hline
    BLEU & 16.37 & 16.54 & \textbf{16.59} & 16.54 & 16.02 \\
    \hline
    \end{tabular}
    \caption{Different alpha values result.}
\end{table}

\subsection{Hyperparameters}
\begin{table}[h]
    \centering
    \begin{tabular}{|c|cc|}
        \hline
        Parameter & Baseline & Ours \\ 
        \hline
        model & BART-base & BART-base \\
        training epochs & 5 & 5 or 30 \\
        learning rate & 6.25-e5 & 6.25-e5 \\
        training batch size & 2 & 2 \\
        \textbf{alpha (length norm)} & 0.0 & 1.0 \\
        \textbf{beam size} & N/A (nucleus) & 10 \\
        \textbf{minimum length} & 1 & 5 \\
        \textbf{maximum length} & 20 & 80 \\
        \hline
    \end{tabular}
    \caption{Hyperparameters of the baseline model and ours. Parameters with bold represent decoding hyperparameters.}
    \label{hyperparameters}
\end{table}
\end{document}

%% file: introduction.tex
\section{Introduction}

Call For Customized Conversation~\cite{pkchat2022} is an open-domain chat dataset that grounds human dialogue in both Persona and Knowledge. The dataset provides Knowledge grounded multi-turn dialogue that are aligned with user's Persona. In particular, this dataset explores how variety of people's individual preferences affects required Knowledge selection to generate the answer while travelling around the world (history, design, structure, tourism etc.). Thus, the dataset is composed of dialogues annotated with individual landmark associated Wiki passages and simple sentences inferring user's preferences. This results in more realistic dialogue environment for evaluation of open-domain dialogue agents.

One important aspect of this configuration is that Persona and Knowledge pairs should be retrieved from given dialogue. Following grounding prediction tasks in~\cite{pkchat2022}, we define \textit{Persona and Knowledge Dual Context Identification} as the task to identify Persona and Knowledge jointly for a given dialogue. We hypothesize that there are specific interactions that happen between Persona, Knowledge, and Dialogue, thus they cannot be predicted separately from partial contexts. We utilize neural retrieval tools such as Sentence-BERT~\cite{sbert2019} to jointly predict Persona and Knowledge. This is the first paper to outline joint retrieval techniques for multi-context grounded dialogue as of our understanding.

In addition, decoding techniques are crucial since conversation models have same encoder-decoder architecture utilized for other Text Generation tasks. \cite{roller2020recipes} introduce recipes for retrieval and generation models where they emphasize decoding choices for grounded open-domain dialogue. \cite{wu2016google} propose a variety of normalization techniques for machine translation in a production system (Google Translate). \cite{meister2020if} investigates importance of beam configurations in reaching optimal performance. Following these studies, we aim to tackle known problem of brevity in which generative models favor shorter, less informative text than is optimal. We extensively experiment with various decoding strategies, length constraints and normalization techniques.

Our contributions are as follows :

1. Persona-Knowledge dual context retrieval methodology which utilizes neural retrieval tools to jointly retrieve Persona and Knowledge given Dialogue. We achieve SOTA performance for both Persona and Knowledge retrieval. Notably, no model fine-tuning is required for top-1 Knowledge retrieval method.

2. Enhanced decoding strategy that target optimal performance with specific emphasis on brevity enhancement. Notably, our approach obtains a significant performance gain without additional data or training.

%% file: related.tex

\section{Related Works}

Integrating Persona with dialogue agents has been actively studied. Various different datasets and systems exist for the purpose, including Persona Chat~\cite{zhang-etal-2018-personalizing} and many others~\cite{majumder-etal-2020-like, persona-goal, persona-image, persona-topical, persona-towards}. Access to Persona assists the dialog agent in responding correct dialogue to the user, however, lack of Knowledge context prohibits the agent from elaborating with specific detailed information.

On the other hand, integrating knowledge bases with dialogue is another engaging topic of dialogue studies. Datasets for this purpose are~\cite{dinan2018wizard, zhou2018dataset}. Relevant Knowledge to the dialogue is retrieved from the knowledge base and utilized in response generation. The shortcoming of this Knowledge-only approach is that relevant Knowledge itself might depend on Persona of the user. We specifically address this shortcoming in our method via studying interactions between all components of dialogue. 

In dialogue generation, \cite{wu2016google} propose a variety of beam normalization techniques for machine translation. \cite{roller2020recipes} emphasizes decoding strategies for open-domain chatbot including beam size, beam length, and sampling methods. \cite{meister2020if} introduces regularization strategies for beam search.


%% file: methodology.tex
\section{Methodology}

\subsection{Knowledge Retrieval}
\label{kr_subsection}

We introduce a novel formulation of Persona, Knowledge and Dialogue as Q \& A input (Figure \ref{QAform}). This form is specifically selected to infer relations between all inputs of the grounded dialogue during answer likelihood calculation, and to replicate short question and descriptive answer pairs often found in Q \& A setting. $E$ notates pair for inference with retrieval model, $Q_i, A_j$ notates specific Q \& A candidate pairs, $P_i, K_j$ notates specific Persona and Knowledge pairs respectively, and $D$ notates dialogue corresponding to the pairs.

\begin{equation}
E : \{Q_i, A_j\} = \{P_i + D, K_j\}
\label{QAFormEq}
\end{equation}

\begin{figure}
\begin{tikzpicture}
\node[draw, align=left] at (0,0) { Question : "\{I want to visit Seven Wonders of the\\ Ancient World.\} \{Wow, what is this?\}" \\ \\ Answer : "\{The Great Pyramid of Giza ... of the\\ Seven Wonders of the Ancient World, ...\}"};
\end{tikzpicture}
\caption{Q\&A formulation of Persona \& Knowledge pair (eq. \ref{QAFormEq}). Question form is "\{Persona\} \{Dialogue\}" while answer is "\{Knowledge\}".}
\label{QAform}
\end{figure}

\begin{figure}
\begin{tikzpicture}[auto,
    node distance = 5mm and 24mm,
    centered
    ]
\hspace{16mm}
\node(n11)[circle, draw] {P 1};
\node(n21)[circle, draw, below=of n11] {\textbf{P 2}};
\node(n31)[circle, draw, below=of n21] {P 3};
\node(n41)[circle, draw, below=of n31] {P 4};
\node(n51)[circle, draw, below=of n41] {P 5};
\node(n12)[circle, draw, right=of n11] {K 1};
\node(n22)[circle, draw, right=of n21] {K 2};
\node(n32)[circle, draw, right=of n31] {\textbf{K 3}};
\node(n42)[circle, draw, right=of n41] {K 4};
\node(n52)[circle, draw, right=of n51] {K 5};
\draw (n11) edge [dashed] (n12)
    (n11) edge [dashed] (n22)
    (n11) edge [color=red] (n32)
    (n11) edge [dashed] (n42)
    (n11) edge [dashed] (n52);
\draw (n21) edge [dashed] (n12)
    (n21) edge [dashed] (n22)
    (n21) edge [color=red, line width=0.5mm] (n32)
    (n21) edge [dashed] (n42)
    (n21) edge [dashed] (n52);
\draw (n31) edge [dashed] (n12)
    (n31) edge [dashed] (n22)
    (n31) edge [color=red] (n32)
    (n31) edge [dashed] (n42)
    (n31) edge [dashed] (n52);
\draw (n41) edge [dashed] (n12)
    (n41) edge [dashed] (n22)
    (n41) edge [color=red] (n32)
    (n41) edge [dashed] (n42)
    (n41) edge [dashed] (n52);
\draw (n51) edge [dashed] (n12)
    (n51) edge [dashed] (n22)
    (n51) edge [color=red] (n32)
    (n51) edge [dashed] (n42)
    (n51) edge [dashed] (n52);
\end{tikzpicture}
\caption{Persona-Knowledge permutations computed in search for best Persona \& Knowledge. Best Persona \& Knowledge pair is P2 and K3. Candidate pairs for Persona search (eq. \ref{KtrueInput}) are marked in red. }
\label{PKPermuteDiagram}
\end{figure}
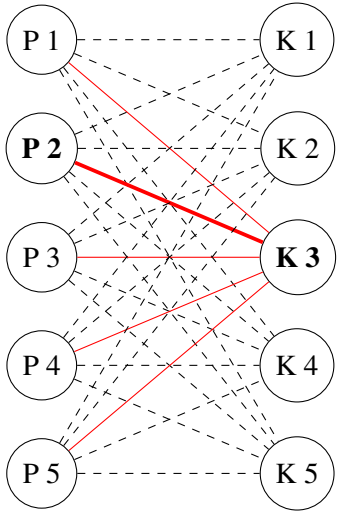

We then perform permutative Persona-Knowledge evaluation (Figure \ref{PKPermuteDiagram}) on all pairs of augmented Persona and Knowledge $E$. We find the best Knowledge via computing all pairs and recording Knowledge of most aligned pair. This is to make sure we find the best Knowledge that aligns with the Dialogue and Persona of the human. $M_q$ notates Q \& A retrieval model that returns relevancy score and $true_j$ notates index of predicted true Knowledge $K$.

\begin{equation}
\begin{aligned}
true_j = \argmax_j M_q\{P_i + D, K_j\} \\
\text{for }i \in {1...n}, j \in {1...m}.
\end{aligned}
\label{KtrueModel}
\end{equation}
\subsection{Persona Retrieval}

Continuing from Section \ref{kr_subsection}, we fine-tune the Q \& A retrieval model $M_q$ using augmented Persona and predicted true Knowledge $K_{true_j}$ pairs only, without incorrect Knowledge pairs. This fine-tuning step is to increase the performance of the model, and obtain correct normalized scores for Persona. Otherwise we will obtain higher scores due to alignment of $D$ with $K$ in terms of Q \& A configuration. $M_f$ notates the fine-tuned model. $E'$ is input to the Q \& A model similar to $E$, only difference being fixed true Knowledge $K_{true_j}$. We note $E'_{train}$ separately because it is data from separate training set formulated in same manner as $E'$ with labeled true Knowledge.

\begin{equation}
E' : \{Q_i, A_{true}\} = \{P_i + D, K_{true_j}\}
\label{KtrueInput}
\end{equation}

\begin{equation}
M_q \xrightarrow{E'_{train}} M_f
\label{Finetune}
\end{equation}

Finally, we infer $E'$ data pairs with model $M_f$ to obtain Persona likelihood score. We utilize a threshold $p_{thres}$ to avoid retrieving unrelated Persona. Certain Dialogue has no Persona assigned to it, which we can replicate with the threshold.

\begin{equation}
p_i = M_f\{P_i + D, K_{true_j}\}
\label{Palign}
\end{equation}

\begin{equation}
\begin{aligned}
true_i = \argmax_i
\begin{cases}
     p_i,& \text{if } p_i \geq p_{thres} \\
    0,& \text{otherwise}
\end{cases}\\
\text{for }i \in {1...n}.
\end{aligned}
\label{PersonaEq}
\end{equation}

Retrieved Persona and Knowledge for given Dialogue $D$ is as follows, notated by $R$ :

\begin{equation}
R : \{D, P, K\} = \{D, P_{true_i}, K_{true_j}\}
\label{Palign}
\end{equation}

\subsection{Decoding Techniques}

We describe generated grounded conversation response as a downstream task of Persona-Knowledge retrieval.

\begin{equation}
    S = G\{D, P, K, L_{min}, L_{max}, \alpha , \beta \}
\end{equation}
where $S$ indicates a response, $D$ represents a dialogue, $P$ denotes persona, $K$ indicates knowledge, $L_{min}$ represents minimum response length, $L_{max}$ denotes maximum response length, $\alpha$ indicates the coefficient of the length normalization, $\beta$ denotes a beam size and $G$ represents the dialogue generation model. Note that we utilize our implementation of beam search instead of nucleus sampling \cite{nucleus} baseline from \cite{pkchat2022}.\\
\indent For length normalization technique, we apply the following formula proposed by~\cite{wu2016google} to our decoder with various alpha values. We report experimental result in Appendix \ref{sec:appendix}.
\begin{equation}
    length\_norm(Y) = \frac{(5+|Y|)^\alpha}{(5+1)^\alpha}
\end{equation}
where |Y| denotes the current target length and $\alpha$ indicates the length normalization coefficient.

%% file: experiment_setup.tex
\section{Experiment Setup}

\label{experiments}

We utilize Call For Customized Conversation~\cite{pkchat2022} dataset for evaluation and fine-tuning, which has 10 Knowledge and 5 Persona candidates respectably for each dialogue. We integrate neural Question and Answering retrieval model from Sentence-BERT~\cite{sbert2019} as starting model $M_q$. Specifically, we utilize 12 layer MiniLM~\cite{wang2020minilm} (33M params) based cross-encoder trained on MS MARCO\footnote{MRR@10 on MS MARCO Dev Set: 39.02}~\cite{nguyen2016ms}. This model fits very well with our formulation since its purpose is for semantic search, with model evaluating short questions and long passages together. For Persona search (eq. \ref{Finetune}, \ref{PersonaEq}), we fine-tune for 2 epochs and provide threshold of $0.5$ in our best configuration.\\ \indent In addition, to evaluate generation task, we extensively experiment with baseline generation model trained via configuration in~\cite{pkchat2022} combined with several decoding hyperparameters. We train the baseline model for 5 epochs, and we use default decoding settings as minimum length 1, maximum length 20, and nucleus sampling. Finally, our method is trained additionally 25 epochs and uses minimum length 5, maximum length 80, and beam size 1, alpha 1.0. Exact hyperparameters are attached to Table~\ref{hyperparameters}.

%% file: results.tex
\section{Results}

\subsection{Knowledge Retrieval}
\label{kr_section}

We experiment with various ablations of Dialogue / Persona / Knowledge interactions and find permutative evaluation of eq.\ref{QAFormEq} form yields best performance for selecting top-1 Knowledge. Result of 15 point increase confirms that considering all components of dialogue is important. We report the results on test set.

\begin{table}[h]
    \centering
    \begin{tabular}{|c|c|}
        \hline
        Model Type & Accuracy \\
        \hline
        D \& K & 79.26 \\
        P \& K (pairwise) & 84.62 \\
        P + D \& K (pairwise) & \textbf{94.69 (+15.41)} \\
        \hline
    \end{tabular}
    \caption{Knowledge retrieval results.}
\end{table}


\subsection{Persona Retrieval}

For Persona retrieval experiments, we start with grounding Knowledge $K_{true}$ selected in Section \ref{kr_section}. Then, we perform ablations of Dialogue augmentation and fine-tuning. Fine-tuning of $P + D$ model yields 8 point performance increase.

\begin{table}[h]
    \centering
    \begin{tabular}{|c|c|}
        \hline
        Model Type & Accuracy \\
        \hline
        P \& $K_{true}$ & 86.75 \\
        P + D \& $K_{true}$ & 83.83 \\
        P + D \& $K_{true}$ (fine-tuned) & \textbf{91.57 (+7.74)} \\
        \hline
    
    \end{tabular}
    \caption{Persona retrieval results with threshold 0.5.}
\end{table}

We observe low performance for $P + D$ in comparison to $P$. We suspect that this is due to lack of score normalization, in that Q \& A relationship of Dialogue to true Knowledge may affect likelihood score. We argue that fine-tuning $P + D$ model normalizes the score in addition to raw performance increase. We perform threshold ablations as shown in Table~\ref{persona_ablation} to verify our hypothesis.

\begin{table}[h]
    \centering
    \begin{tabular}{|c|c|c|c|c|}
        \hline
        Model Type & 0.0 & 0.5 & 0.6 & 0.7 \\
        \hline
        P + D \& $K_{true}$ & 79.30 & 83.83 & 84.02 & \textbf{84.26}\\
        Fine-tuned & 86.81 & 91.57 & \textbf{92.16} & 91.87\\
        \hline
    \end{tabular}
    \caption{Persona retrieval threshold ablations.}
    \label{persona_ablation}
\end{table}

We find that fine-tuned model has increased performance across all thresholds, including $0.0$ where the output has top-1 characteristics. We also find that the score increases in tandem with Persona threshold for non-fine-tuned case.

\subsection{Generation Results}

We experiment with various decoding methods and perform ablations. In these experiments, we use ours (5 epoch) model described in Table~\ref{hyperparameters}. We report the results on dev set.

\textbf{
\begin{itemize}
    \item Q1. What is the optimal performance we can reach with decoding method improvements?
    \item Q2. How does the decoding strategy affect performance?
    \item Q3. How does the length constraints affect performance?
\end{itemize}
}

\noindent\textbf{Q1. What is the optimal performance we can reach with decoding method improvements?}\\
We obtain 10, 11 point increase of BLEU and Rough-L respectably as described in Table~\ref{overall}.

\begin{table}[h]
    \centering
    \begin{tabular}[width=0.5]{|c|cc|}
    \hline
     Model & Rouge-L & BLEU\\
     \hline
     Baseline & 30.79 & 11.16 \\
     \hline
     Ours (5 epoch) & 38.50 (+7.71) & 19.31 (+8.15)\\
     \hline
     Ours (30 epoch) & \textbf{41.54 (+10.75)} & \textbf{21.42 (+10.26)} \\
     \hline
    \end{tabular}  
    \caption{Performance report of our method.}
    \label{overall}
\end{table}

\noindent\textbf{Q2. How does the decoding strategy affect performance?}\\
We select beam size of 10 informed by~\cite{meister2020if}. Table~\ref{beamsize} demonstrates effectiveness of beam search compared to baseline nucleus sampling.

\begin{table}[h]
    \centering
    \begin{tabular}{|c|cc|}
    \hline
     Beam Size & N/A (nucleus) & 10\\
     \hline
     BLEU & 13.76 & \textbf{19.31 (+5.55)} \\
     \hline
    \end{tabular}  
    \caption{BLEU score of ours (5 epoch) model with different sampling strategies.}
    \label{beamsize}
\end{table}

\noindent\textbf{Q3. How does length constraints affect performance?}\\
Table~\ref{maximum} demonstrates that the longer the maximum response length, the higher the performance gain. We also experiment with minimum length constraints in Appendix~\ref{sec:appendix}.
 \begin{table}[h]
    \centering
    \begin{tabular}{|c|cccc|}
    \hline
     Max. Length & 20 & 40 & 60 & 80\\
     \hline
     BLEU & 12.86 & 16.37 & 17.20 & \textbf{19.31} \\
     \hline
    \end{tabular}
    \caption{BLEU score of ours (5 epoch) model with different maximum lengths.}
    \label{maximum}
\end{table}

%% file: conclusion.tex
\section{Conclusion}

We introduce Persona-Knowledge dual context retrieval method in this paper. We achieve SOTA grounding retrieval performance by Q \& A informed data augmentations and application of novel fine-tuning techniques. We achieve SOTA dialogue generation performance by utilizing beam search and brevity-informed constraints. We perform minimal fine-tuning for both high-performing methods. We are first place across all metrics (Persona / Knowledge accuracy, SacreBLEU, CharF++, ROUGE-L) in the official leaderboard. We achieve significant $10+$ point increase over each baseline metrics for both Grounding and Generation tasks.